\pdfoutput=1

\documentclass[11pt]{article}

\usepackage[final]{acl}
\usepackage{listings}
\usepackage{times}
\usepackage{latexsym}
\usepackage{booktabs}
\usepackage[utf8]{inputenc}
\usepackage{xcolor}
\usepackage{boxedminipage}
\usepackage{subfig}

\newcommand{\dart}{DART\includegraphics[height=.7em]{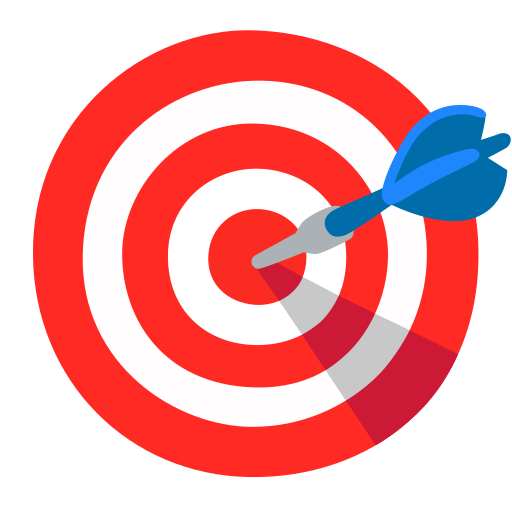}}
\newcommand{\dartdt}{DART\textsubscript{DT}}
\newcommand{\dartsvm}{DART\textsubscript{SVM}}

\newcommand{\currentlywriting}{}

\usepackage[T1]{fontenc}

\usepackage[utf8]{inputenc}

\usepackage{microtype}

\usepackage{inconsolata}

\usepackage{graphicx}

\title{\dart: An AIGT Detector using AMR of Rephrased Text}

\author{Hyeonchu Park\textsuperscript{\textdagger}, Byungjun Kim\textsuperscript{\textdagger}, and Bugeun Kim
\\
        Department of Artificial Intelligence, Chung-Ang University, Republic of Korea \\
        \{phchu0429, k36769, bgnkim\}@cau.ac.kr}

\begin{document}
\maketitle
\def\thefootnote{\textdagger}\footnotetext{Equal contribution.}\def\thefootnote{\arabic{footnote}}
\begin{abstract}
As large language models (LLMs) generate more human-like texts, concerns about the side effects of AI-generated texts (AIGT) have grown. So, researchers have developed methods for detecting AIGT. 
However, two challenges remain. 
First, the performance of detecting black-box LLMs is low because existing models focus on probabilistic features.
Second, most AIGT detectors have been tested on a single-candidate setting, which assumes that we know the origin of an AIGT and which may deviate from the real-world scenario.
To resolve these challenges, we propose \dart, which consists of four steps: rephrasing, semantic parsing, scoring, and multiclass classification.
We conducted three experiments to test the performance of DART.
The experimental result shows that DART can discriminate multiple black-box LLMs without probabilistic features and the origin of AIGT. 
\end{abstract}

\section{Introduction}
As large language models (LLMs) continue to advance, it becomes increasingly difficult for humans to discern AI-generated text (AIGT). This poses issues in society and research, such as spreading fake news and tainting AI training data. Researchers have developed AIGT detectors to address these issues. Despite their success, two challenges related to real-world applicability persist.


One challenge with applying AIGT detectors is low performance in detecting recent black-box LLMs. Traditionally, AIGT detectors rely on probabilistic features such as logits. However, in commercial black-box models, including GPT \cite{gpt4o, gpt3.5} or Gemini \cite{gemini}, we cannot access their computation procedure which provides logits. That is, traditional approaches cannot detect such black-box models. So, researchers have also designed detectors using syntactic features that do not require accessing computational procedures. Yet, these detectors struggle to detect black-box models because their syntactic naturalness is comparable to that of humans. 


The other challenge is the vagueness of the origin of AIGTs. In the inference time of a detector, it receives a text without any information about its origin. So, similar to the inference scenario, we should verify whether a detector can successfully discriminate AIGT regardless of source models. However, existing studies mainly tested their detectors under the assumption that a candidate LLM is known in advance; they tested whether a binary detector can distinguish a `human-written text' from an `AIGT by the predefined candidate.' 
So, whether existing detectors can detect the origin without the assumption is questionable.


To address these challenges, we propose a Detector using AMR of Rephrased Text (\dart).
DART utilizes the semantic gap between given input and rephrased text, using Abstract Meaning Representation (AMR). 
This rephrasing idea was first introduced by RAIDAR \citep{raidar}; we adopted a similar idea to reveal such a semantic gap.
To examine the real-world detection performance, we assess DART in three settings: single-candidate, multi-candidate, and leave-one-out. Experimental results show that DART can successfully discriminate humans from four cutting-edge LLMs, including GPT-3.5-turbo, GPT-4o, Llama 3-70b \cite{llama}, and Gemini-1.5-Flash.

Thus, this paper has the following contributions:
\begin{itemize}
    \item We present a semantics-based detection framework for AIGT, leveraging semantic gaps between given input text and rephrased texts.
    \item DART can discriminate different LLMs and outperform other models. On average, DART beat others by more than 19\% in F1 score.
    \item Also, DART can generalize its knowledge on detecting unseen source models. Specifically, DART achieved a 85.6\% F1 score on leave-one-out experiment.
\end{itemize}

\section{Background}





In this section, we categorize existing studies regarding numbers (\textit{single} or \textit{multi}) and transparency (\textit{white box} or \textit{black box}) of candidate LLMs.


\paragraph{Single white-box candidates} AIGT detectors first attempted to extract candidate-specific features. As the candidate is a known white-box model, some researchers designed algorithms adopting probabilistic features from the model \citep{gltr, detectgpt}. For example, DetectGPT \citep{detectgpt} used log probabilities of tokens as features. Other researchers used neural models that can learn features from the given texts \citep{roberta, radar}. However, as many black-box LLMs recently emerged, the performance of existing detectors should be revalidated on those LLMs.







\paragraph{Single black-box candidates} Some AIGT detectors then attempted to extract features regardless of the candidate \citep{fast_detect, dpic, dnagpt, motif}, as black-box candidates may not provide probabilistic features. Fast-DetectGPT \citep{fast_detect} extended DetectGPT by extracting probabilistic features from a proxy white-box model (e.g., GPT-J). Since such a proxy can provide less accurate results, other studies used syntactic or surface-level features without using a proxy \citep{dnagpt, motif}. For example, DNA-GPT \citep{dnagpt} used $n$-grams from multiple paraphrased texts generated by the candidate. However, such syntactic features are insufficient to detect recent LLMs because recent models generate text with human-level syntax.



\paragraph{Multiple candidates} As a single-candidate performance is far from real-world scenarios, recent AIGT detectors were designed to detect multiple candidates \citep{sniffer, abburi, seqxgpt, poger, antoun}. For example, POGER \citep{poger} extends resampling methods to estimate probability using about 100 paraphrases. Because of such an excessive regeneration, POGER incurs high computational costs. Besides, SeqXGPT \citep{seqxgpt} used a Transformer-based detector with a proxy model estimating probabilistic features. However, these studies mainly focused on surface-level features and the limited range of LLMs (e.g., GPT family), raising questions about detecting other cutting-edge LLMs.

\section{The DART Framework}
\label{sec:dart}

\begin{figure}[t]
  \includegraphics[width=.95\columnwidth]{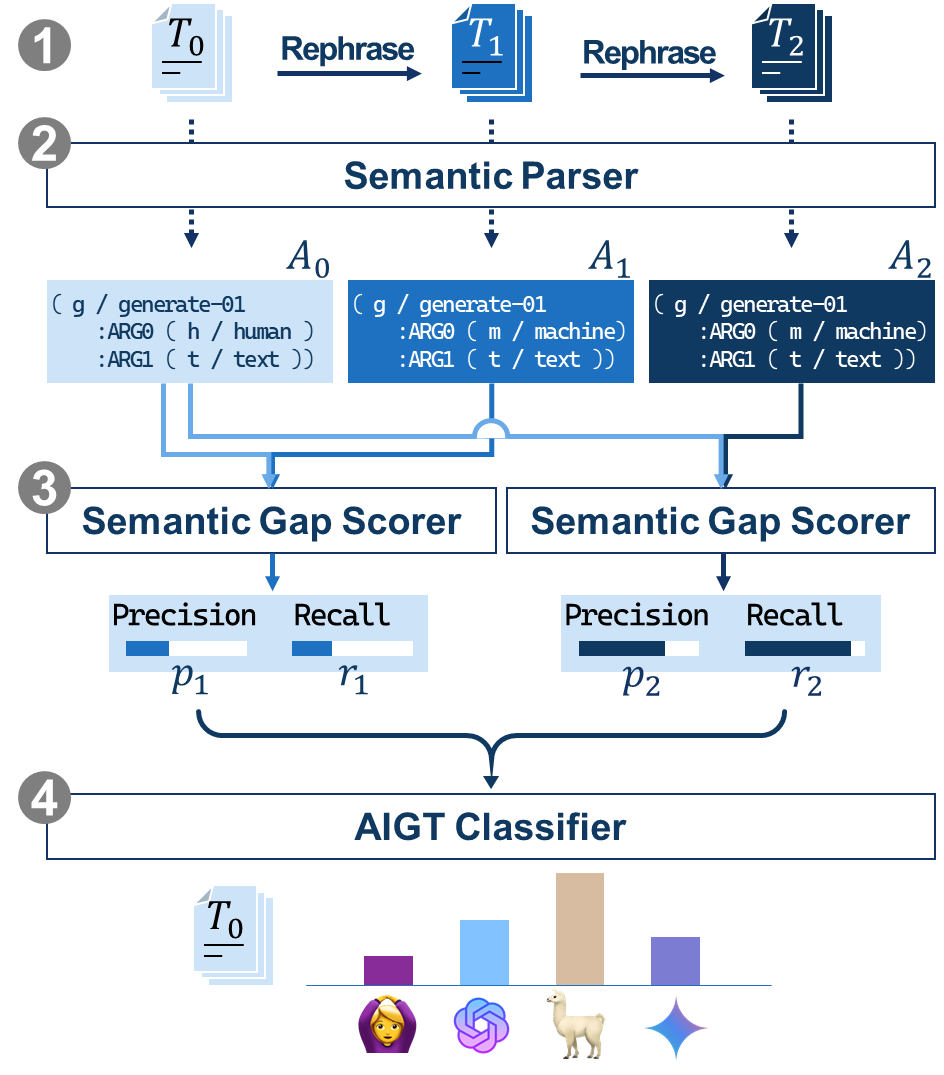}
  \caption{The DART framework}
  \label{fig:flowchart}
\end{figure}

As shown in Figure \ref{fig:flowchart}, DART utilizes semantic gaps between a given text and rephrased texts. To train a detector capturing such gaps, DART uses a four-step procedure: \textit{Rephrasing}, \textit{Semantic parsing}, \textit{Semantic gap scoring}, and \textit{Classification}.

\paragraph{Step 1, Rephrasing:}
We hypothesized that rephrasing texts could reveal the difference between humans and AI in the way they express semantics. To obtain the rephrased texts, DART uses a \textit{rephraser} module that generates semantically closer text $T_1$ from a given text $T_0$.
Further, we let the rephraser generate another rephrased text $T_2$ by giving $T_1$ to attain additional features. 
To avoid generating rephrased texts irrelevant to the given input, we need a reliable rephraser that can preserve semantics. So, we adopted \texttt{GPT-4o-20240513} as our rephraser because the model showed the highest performance in semantics-related tasks \cite{gpt4o}.
Appendix \ref{app:rephraserprompt} details the prompts used in the rephrasing step.



\paragraph{Step 2, Semantic parsing:} DART adopts a semantic parser to transform texts into semantic representations. We especially adopted AMR as a semantic representation because AMR has widely been adopted to abstract the given text into semantics \citep{AMR}. For the parser, we adopted \citet{DocAMR}. As a result, the parser constructs an AMR graph $A_i$ from each $T_i$. 

\paragraph{Step 3, Semantic gap scoring:} DART uses metrics for semantic parsers to measure semantic gaps between texts. As we adopted AMR as a semantic representation in the previous step, we utilize a fast and efficient algorithm for scoring AMR similarity called SEMA \citep{sema, ki}. To obtain semantic gaps between $A_0$ and $A_i$ ($i>0$), DART computes precision $p_i$ and recall $r_i$ scores generated by SEMA, resulting a feature vector $v = [p_1, p_2, r_1, r_2]^\top$ for the next step.

\paragraph{Step 4, Classification:} DART has a classifier that predicts one possible origin of $T_0$. DART uses interpretable classifiers, including support vector machine (SVM) or decision tree (DT), though any classifier that maps $v$ to origins can be used.

\section{Experiments}
To evaluate the performance of DART, we conducted three experiments: (1) single-candidate, (2) multi-candidate, and (3) leave-one-out settings. First, in the single-candidate setting, we formulate AIGT detection as a binary classification task. Assuming that AIGTs are exclusively produced by a specific LLM, a detector should predict whether the given text is produced by the LLM.
Second, in the multi-candidate setting, we formulate the task as a multi-label classification. After training on AIGTs from multiple candidate sources, a detector should decide the source of the given input text among the candidates.
Third, in the leave-one-out setting, we test the generalizability of detectors. 
We examined whether a detector can successfully classify AIGTs from models that were unseen during the training.

We ran each experiment 10 times for each experiment to achieve reproducibility.
Further, we analyzed DART's training efficiency by examining the decreasing rate of detecting performance as the size of the training dataset.

\subsection{Datasets}
To train DART, we need human-written texts and AIGTs. First, we used four English datasets as human-written text datasets: XSum \cite{xsum}, SQuAD 1.1 \cite{squad}, Reddit \cite{reddit}, and PubMedQA \cite{pubmed}. 
Following the practice of previous research \citep{detectgpt, seqxgpt}, we randomly sampled texts from these datasets. We split training and validation sets with an 8:2 ratio.



Second, we generated AIGT datasets based on the human dataset. Following \citet{detectgpt}, we collected English AIGT from each human-written text. Four cutting-edge LLMs are used to generate AIGTs: GPT-4o, GPT-3.5-turbo, Llama 3-70B, and Gemini-1.5 Flash. We obtained AIGTs by providing the first 30 tokens of each human-written text to an LLM. Because PubMedQA contains many texts shorter than 30 tokens, we provided corresponding questions instead of the first 30 tokens. Appendix \ref{app:aigtprompt} illustrates the detailed prompts used for generating AIGTs.
%
%
As a result, we obtained about 3,989 human-written texts and 15,956 AIGTs (= 3,989 texts $\times$ 4 LLMs). 
See Appendix \ref{app:statistics} for the statistics of the collected dataset.

\subsection{Baselines}

As baselines, we used five open-source state-of-the-art detectors: DetectGPT \cite{detectgpt}, Fast-DetectGPT \cite{fast_detect}, DNA-GPT \cite{dnagpt}, Roberta-base \cite{roberta}, and SeqXGPT \cite{seqxgpt}. Among these models, DetectGPT, Fast-DetectGPT, and SeqXGPT used probabilistic features generated by third-party LLMs in order to detect cutting-edge LLMs. Meanwhile, DNA-GPT and Roberta-base used shallow semantic features, such as $n$-grams or contextual embeddings. 
DART stands out from these models because it uses AMR-based semantics rather than probabilistic features.

We used a different set of detectors for the three experiments, considering experiments reported with five baselines. For the single-candidate experiment, we compared DART with all five detectors. For the multi-candidate and the leave-one-out experiments, we compared DART only with SeqXGPT, as it is the only existing detector that can trained on multiple candidates simultaneously. To ensure a fair comparison, all detectors used in the experiment are trained on our dataset from scratch\footnote{Note that we used GPT-2 as a proxy model for the GPT series and Gemini-1.5 when the detectors require probabilistic features because GPT and Gemini do not provide logits, following \cite{fast_detect}.}.
%
%
To measure the performance, we used the F1 score.

\begin{table*}[h!] 
\centering
\begin{tabular}{lr|r@{$\pm$}lr@{$\pm$}lr@{$\pm$}lr@{$\pm$}l}
\toprule
                      & Average           & \multicolumn{2}{c}{GPT-3.5-turbo}     & \multicolumn{2}{c}{GPT-4o}            & \multicolumn{2}{c}{Llama3-70B}            & \multicolumn{2}{c}{Gemini-1.5}        \\ 
\midrule
DetectGPT*             & 65.8       & 65.8&0.20    & 65.6&0.16     & 65.8&0.17        & 65.7&1.12          \\
fast-DetectGPT*           & 60.1       & 58.0&1.94    & 66.2&0.25     & 62.4&0.48        & 53.8&0.58          \\
DNA-GPT                & 54.1       & 56.6&1.49    & 57.4&0.50     & 54.8&2.60        & 47.7&2.36          \\
Roberta-base           & 77.2      & 76.8&3.24      & 80.0&2.81     & 74.7&1.77      & 77.1&2.13\\
SeqXGPT*               & 54.1      & 86.5&0.48      & 45.9&0.23     & 41.6&0.31      & 42.3&0.52                                 \\
\midrule
DART\textsubscript{SVM}            & 82.8      &  87.1&0.65   &  86.1&0.70   & 84.8&2.20    &  73.3&0.76          \\
\phantom{DART}\textsubscript{DT}     & \textbf{96.5}     &  \textbf{100.0}&\textbf{0.03} &  \textbf{88.1}&\textbf{0.98} &  \textbf{100.0}&\textbf{0.03} &  \textbf{97.9}&\textbf{1.65} \\ 
\bottomrule
\multicolumn{10}{r}{\small * Models used GPT-2 as a proxy model, except Llama 3.}
\end{tabular}%
\caption{F1 scores of detectors in the \textbf{single-candidate} experiment, with standard deviations reported.}
\label{tab:single-model result2}
\end{table*}

\section{Result and Discussion}
\label{sec: result and discussion}
\paragraph{Single-candidate experiment:} DART outperformed existing models. As shown in Table \ref{tab:single-model result2}, our \dartdt\ and \dartsvm\ achieved 96.5\% and 82.8\% F1 scores on average, which are 19.3\%p and 5.6\%p higher than the Roberta-base model (77.2\%). Also, \dartdt\ can detect all four cutting-edge models with over 85\% of F1 score. Meanwhile, other existing models showed F1 scores lower than 70\%, on average. Moreover, DNA-GPT and SeqXGPT sometimes showed F1 scores lower than the random binary baseline (50\%).

\begin{table*}[h!] 
\centering
\begin{tabular}{lr|ccccc}
\toprule
                      & Macro F1             & {GPT-3.5-turbo}     & {GPT-4o}            & {Llama3-70B}            & {Gemini-1.5}      &{Human}  \\ 
\midrule
SeqXGPT*       & 59.2$\pm$0.66      & 54.3$\pm$1.44      & 66.4$\pm$1.08     & 44.8$\pm$0.95          & 61.4$\pm$1.68         &69.3$\pm$0.93         \\
\midrule
DART\textsubscript{SVM}    & 65.0$\pm$0.77   & 67.4$\pm$0.81    & 71.0$\pm$1.20     & 54.0$\pm$1.26    & 67.0$\pm$0.94    &65.4$\pm$1.16   \\
\phantom{DART}\textsubscript{DT}        & \textbf{81.2$\pm$1.71} &   { \textbf{80.6$\pm$4.61}}   & { \textbf{76.6$\pm$1.16}}     & { \textbf{85.8$\pm$5.42}}         & { \textbf{85.5$\pm$2.36}}          &{\textbf{77.3$\pm$0.88}}\\ 
\bottomrule
\multicolumn{7}{r}{\small * Models used GPT-2 as a proxy model, except Llama 3.}
\end{tabular}%
\caption{F1 scores of detectors in the \textbf{multi-candidate} experiment, with standard deviations reported.}
\label{tab:multi-model result2}
\end{table*}

\begin{table*}[h!] 
\centering

\begin{tabular}{lr|cccc}
\toprule
                      & Macro F1             & {GPT-3.5-turbo}     & {GPT-4o}            & {Llama3-70B}            & {Gemini-1.5}       \\ 
\midrule
SeqXGPT*     &78.5$\pm$1.04    &79.9$\pm$1.39     &80.2$\pm$0.52      &78.8$\pm$0.92     &75.1$\pm$1.31  \\

\midrule
DART\textsubscript{SVM}   &56.3$\pm$0.96    &56.0$\pm$1.31       &59.2$\pm$1.01      &56.4$\pm$0.82           &53.6$\pm$0.70                   \\
\phantom{DART}\textsubscript{DT}        & \textbf{84.2$\pm$1.39} &   { \textbf{99.3$\pm$0.16}}   & { \textbf{75.8$\pm$3.82}}     & { \textbf{99.1$\pm$0.55}}         & { \textbf{62.5$\pm$1.03}}          \\

\bottomrule
\multicolumn{6}{r}{\small * Models used GPT-2 as a proxy model for black-box models, except Llama3}
\end{tabular}

\caption{F1 scores of detectors in the \textbf{leave-one-out} experiment, with standard deviations reported.}
\label{tab:leave-one-out}
\end{table*}

We suspect that \dartdt\ can achieve such outstanding performance because our semantic scoring step can successfully form several clusters according to their origins. To support this argument, we further analyzed the feature vectors of DART using principal component analysis. We found that texts sharing the same source usually form several independent clusters rather than spread over the space. Detailed results are presented in Appendix \ref{app:pca}.

\paragraph{Multi-candidate experiment:} DART also outperformed SeqXGPT. As shown in Table \ref{tab:multi-model result2}, our \dartdt\ and \dartsvm\ achieved 81.2\% and 65.0\% macro F1 scores, which are 22.0\%p and 5.8\%p higher than SeqXGPT (59.2\%). Interestingly, SeqXGPT achieved the lowest F1 score on detecting Llama 3 (44.8\%), but \dartdt\ achieved the lowest score on detecting GPT-4o (76.6\%).

\begin{figure}[t]
  \centering
  \includegraphics[width=.7\columnwidth]{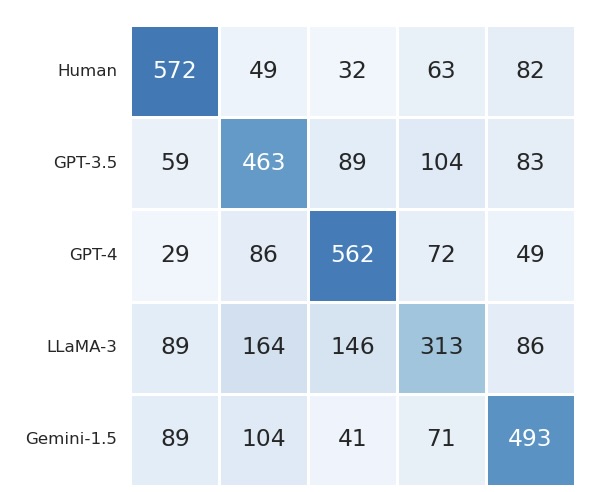}
  \includegraphics[width=.7\columnwidth]{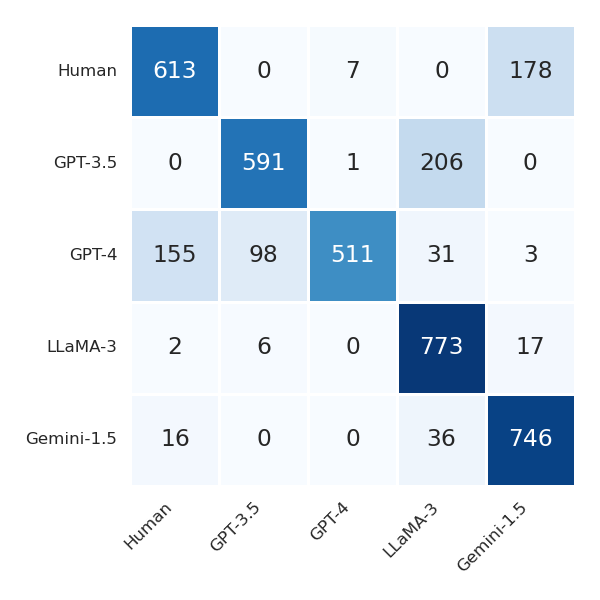}
  \caption {Contingency matrix from multi-candidate experiment. Top (a) and Bottom (b) correspond to SeqXGPT and \dartdt. Actual and predicted classes are depicted as horizontal and vertical axes.}
  \label{fig:contingency}
\end{figure}

We suspect how the detectors extract features using an LLM affects the performance. 
We present a contingency table of SeqXGPT and \dartdt\ to support this claim, 
as shown in Figure \ref{fig:contingency}. The figure shows that (i) SeqXGPT struggled in distinguishing models other than Llama 3, and (ii) \dartdt\ struggled in distinguishing the GPT family and humans. Since SeqXGPT in our experiment used GPT-2 as a proxy model, and \dartdt\ used GPT-4o as a \textit{rephraser} module, the characteristics of the used LLMs affected the detection performances. For example, as \dartdt\ utilizes semantic gaps between the original and rephrased texts, origins should reveal distinguishable gaps to identify them successfully. So, when the gaps are too similar between origins to discriminate them, \dartdt\ faces difficulty in the classification step.

Since GPT-4o has a similar language understanding ability to humans \citep{gpt4o}, GPT-4o and humans may be less distinguishable through gaps. Similarly, as GPT-3.5-turbo may share some core knowledge with GPT-4o, GPT-4o can be confused with GPT-3.5-turbo in \dartdt. 

\paragraph{Leave-One-Out experiment:} \dartdt\ showed the best performance. As shown in Table \ref{tab:leave-one-out}, \dartdt\ achieved 85.6\% average F1 score, followed by SeqXGPT (77.9\%) and \dartsvm\ (56.5\%). Besides, \dartdt\ scored 62.5\% F1 on detecting the unseen Gemini-1.5, though \dartdt\ recorded more than 75\% on detecting others.


This result indicates that \dartdt\ can generalize trained knowledge to detect unseen source models. That is, \dartdt\ can discriminate new candidate models from humans. Specifically, compared to the single-candidate result (Table \ref{tab:single-model result2}), our model showed almost similar performance on detecting GPT-3.5-turbo and Llama 3 without training on those models. As in the single-candidate experiment, we believe that our semantic scoring step helped to detect unseen models because they form clusters independent from humans. Also, when the cluster becomes indiscernible with humans, \dartdt\ struggles to detect new models. For example, \dartdt\ showed a big performance drop when excluding Gemini-1.5 from the training set because \dartdt\ often confused Gemini-1.5 with humans (top-right corner on Figure \ref{fig:contingency}b).

\begin{figure}[t]
  \centering
  \includegraphics[width=1\columnwidth]{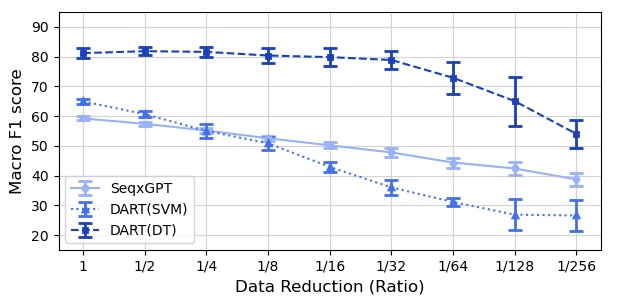}
  \caption{F1 score of detectors when we decrease the amount of training data in multi-candidate experiment.}
  \label{fig:AblationResult}
\end{figure}

\paragraph{Training efficiency of DART:}
Here, we discuss the general tendency of the result. Figure \ref{fig:AblationResult} shows the performance changes when we decrease the size of the training set. Detailed result of training efficiency is presented in Appendix \ref{app:efficiency}. 
The result shows that \dartdt\ is robust even though we use a small amount of training data. Specifically, \dartdt\ maintained a similar F1 score until we used 1/32 of the training set (about 500 examples). Meanwhile, the performance of SeqXGPT and \dartsvm\ monotonically decreases as we reduce the size of the training set.


\currentlywriting
\section{Conclusion}
We introduced an AIGT framework, \dart\, to tackle challenges in applying AIGT detectors to real-world scenarios. 
DART employed \textit{rephraser} and semantic gap scoring module to address the challenges of black-box models.
To evaluate whether DART can address vagueness of origin, we assessed DART in three experimental settings: single-candidate, multi-candidate, and leave-one-out settings. As a result, DART achieved outstanding performance compared to existing AIGT detectors, demonstrating successful capture of differences across origins with semantic gaps.


\section*{Limitations}
Despite the outstanding performance of DART, this paper has three limitations. First, we tested DART only with a single rephraser LLM, GPT-4o. Though GPT-4o provided enough semantic information to distinguish AIGTs successfully, it is questionable whether DART can be used with other rephraser LLMs, such as Llama 3, Gemini Pro, or others. 
Also, we recognize the cost implications of utilizing GPT-4o as a rephraser, which could restrict its applicability in resource-limited environments. Since different language models may provide different rephrased texts with lower costs, we need further study to determine how much rephraser LLM affects the performance. 


Second, the performance of the adopted AMR parser may affect the detection performance of DART. Though the AMR parser rarely introduces errors in the DART framework, such errors may lead to huge changes in detection performance when they occur. Using a publicly available AMR parser \citep{DocAMR}, DART showed the lowest bound of its performance. Thus, we need further study to improve the performance using other semantic parsers.

Third, DART tested on a narrow range of black-box models. While narrow LLMs have become publicly available through paid APIs or pretrained parameters, we tried our best to include recent LLMs, such as Gemini Pro or Claude 3. However, we finally excluded those models because their safeguards prevented from generating AIGTs based on a given human-written text when preparing the AIGT dataset. 
To generalize our findings to other origins, we need to conduct further studies in a broader range of models and design a new method of generating AIGTs.

\section*{Acknowledgments}
This work was supported by Institute of Information \& communications Technology Planning \& Evaluation (IITP) grant funded by the Korea government(MSIT) (No.2021-0-01341, Artificial Intelligence Graduate School Program, Chung-Ang University)

\bibliography{custom}

\newpage
\appendix

\section{Prompts}
\subsection{AIGT datasets}
\label{app:aigtprompt}


    


In general, we followed the prompts used in SeqXGPT \citep{seqxgpt} when generating the AIGT dataset. We collected AIGTs by providing LLMs with the first 30 tokens of human-written texts and letting them generate the rest of the texts, except for the PubMedQA dataset.
Besides, we asked LLMs to answer the questions in the PubMedQA dataset instead of providing the 30 tokens of text, borrowing the collecting method of \citet{detectgpt}. We used different methods for PubMedQA because most of the texts in PubMedQA were shorter than 30 tokens.
In addition, to avoid collecting AIGTs with irrelevant phrases (e.g., ``Here is the generation of ...''), we added a constraint clause in the prompts for Llama 3-70B and Gemini-1.5 Flash.


We understand that different datasets and different prompting methods may affect the performance of the detectors. Therefore, we conducted additional per-subset experiments to investigate whether those differences influenced the detecting performance. The findings are detailed in Appendix \ref{app:prom}.




\paragraph{For GPT family}
When collecting AIGTs with GPT-3.5-turbo and GPT-4o, we used the following prompts except for the PubMedQA dataset.

\begin{boxedminipage}{.9\columnwidth}
\ttfamily\raggedright
Please provide a continuation for the following content to make it coherent: \{first 30 tokens\}
\end{boxedminipage}\\\vspace{1em}

For PubMedQA, we used the following prompts:

\begin{boxedminipage}{.9\columnwidth}
\ttfamily\raggedright
Please answer the question: \{question\}
\end{boxedminipage}

\paragraph{For Llama 3-70B and Gemini-1.5-Flash}
When collecting AIGTs with Llama 3-70B and Gemini-1.5-Flash, we used the following prompts except for the PubMedQA dataset.

\begin{boxedminipage}{.9\columnwidth}
\ttfamily\raggedright
Please provide a continuation for the following content to make it coherent: \{first 30 tokens\}\\
Provide the continuation without any prefix.\\
------\\
answer:
\end{boxedminipage}\\\vspace{1em}
\phantom{a}
\\\vspace{1em}

For PubMedQA, we used the following prompts:

\begin{boxedminipage}{.9\columnwidth}
\ttfamily\raggedright
Please answer the question: \{question\}\\
Provide the continuation without any prefix.\\
------\\
answer:
\end{boxedminipage} \\

\subsection{DART's \textit{rephraser}}
\label{app:rephraserprompt}
When rephrasing a text into another rephrased version, we used the following prompt in the rephraser module.

\begin{boxedminipage}{.9\columnwidth}
\ttfamily\raggedright
Please rewrite the following paragraph in \{n\} words: \{paragraph\}
\end{boxedminipage}\\\vspace{1em}

We used this prompt because we observed some semantic meanings of rephrased texts were largely changed without any prompting method in our pre-experiment. For example, some rephrased texts were much longer or shorter than the original texts, which was enough to distort the core message of the origins.
As such distortion leads to unintended trivial semantic differences, we wanted to avoid such too-short or too-long texts. Thus, we restricted the word counts of rephrased texts by using prompts.
Table \ref{tab:word_count} on page \pageref{tab:word_count} shows the average number of words in the original and rephrased texts that we collected. It shows that the number of words slightly changed after rephrasing. We believe that such changes are minor to affect the performance of DART.

\begin{table}
\centering
\begin{tabular}{lrrr}
\toprule

& \multicolumn{1}{c}{$T_0$}    & \multicolumn{1}{c}{$T_1$}    &\multicolumn{1}{c}{$T_2$}     \\
\midrule
Human      & 267.95 & 258.47 & 270.38 \\
GPT-3.5-T  & 107.48 & 89.85 & 83.08 \\
GPT-4o     & 260.03 & 253.59 & 262.56 \\
Llama3     & 152.33 & 133.94 & 127.69 \\
Gemini-1.5 & 131.32 & 116.74 & 110.25 \\
\bottomrule

\end{tabular}
\caption{Average number of words after rephrasing}
\label{tab:word_count}
\end{table}

\begin{table} 
\centering

\begin{tabular}{@{\;}l@{\;\;}r|c@{\;\;}r@{\;\;}c@{\;\;}c@{\;}}
\toprule
                      & Mac F1             & {\small Xsum}     & {\small SQuad}            & {\small Reddit}            & {\small PubMed}       \\ 
\midrule
SeqXGPT*     &63.0    &75.1   &57.0   &58.2   &61.7  \\

\midrule
DART\textsubscript{SVM}   &88.8    &80.0   &92.4   &93.2        &89.8              \\
\phantom{DART}\textsubscript{DT}      & 98.6  &  99.0  & 98.4   & 98.6  &98.4       \\

\bottomrule
\end{tabular}

\caption{Performance of AIGT detectors across different subsets in a Multi-Candidate setting}
\label{tab:diff_data}
\end{table}

\section{Experimental setting}
\subsection{Environment}
\paragraph{Hardware configuration:} The experiments were conducted on a system with an AMD Ryzen Threadripper 3960X 24-Core Processor and four NVIDIA RTX A6000 GPUs. The four NVIDIA RTX A6000 GPUs are used to train existing detectors and execute AMR parsers. The semantic gap scoring module was run on a single core of the CPU.

\paragraph{LLM APIs:} We used commercial APIs of LLMs to collect AIGTs and rephrased texts. GPT models are called with OpenAI's official API. Llama 3-70B is called with a free API provided by \url{groq.com}. Lastly, Gemini-1.5-Flash is called with OpenRouter's API.


\paragraph{Implementation} We used Python 3.11.7 for implementing \dart. Using \texttt{scikit-learn} library, we implemented \dartsvm\ and \dartdt\ with \texttt{SVC} and \texttt{DecisionTreeClassifier}. We mostly used the basic settings of those classes without conducting a hyperparameter search. The only exception is the depth of the pruned tree in \dartdt, and we set it as 5 based on our heuristic.

\subsection{Dataset statistics}
\label{app:statistics}

Table \ref{tab:statistics} in page \pageref{tab:statistics} shows the statistics of the collected dataset.
We used four datasets, which belong to different domains: Xsum \cite{xsum}, SQuAD \cite{squad}, Reddit \cite{reddit}, and PubMedQA \cite{pubmed}. Xsum is a dataset of news articles and summaries. SQuAD is a question-answering dataset whose questions are based on Wikipedia articles. Reddit is a dataset of human-written stories with writing prompts. PubMedQA is a question-answering dataset on a specialized medical domain.

The statistics show that the average lengths of texts in each dataset are different. For example, Gemini-1.5 usually generates long texts on the PubMedQA dataset, while the model generates short texts on the Xsum and Reddit datasets. On average, it seems that the length of a given text is not a significant factor for discriminating origin.

\section{Additional analysis}

\subsection{Precision, Recall}
\begin{table}
\begin{tabular}{lcccc}
\toprule
     & \multicolumn{1}{c}{$p_1$} & \multicolumn{1}{c}{$p_2$} & \multicolumn{1}{c}{$r_1$} & \multicolumn{1}{c}{$r_2$} \\
\midrule
Human      & 0.619 & 0.582 & 0.600 & 0.561 \\
GPT-3.5-T  & 0.645 & 0.605 & 0.631 & 0.595 \\
GPT-4o     & 0.636 & 0.596 & 0.623 & 0.587 \\
Llama3     & 0.648 & 0.610 & 0.631 & 0.594 \\
Gemini-1.5 & 0.651 & 0.615 & 0.633 & 0.596 \\
\bottomrule

\end{tabular}
\caption{Precision and Recall values for text comparisons between {$T_0$}, {$T_1$} and {$T_0$}, {$T_2$}}
\label{tab:pr_diff}
\end{table}

As we discussed in Section \ref{sec:dart}, DART computes precision $p$ and recall $r$ scores with SEMA. Note that $p_i$ and $r_i$ refer to the semantic similarity between the original text $T_0$ and the $i$-th rephrased text $T_i$. DART assumes that the differences between those rephrased texts in terms of $p$ and $r$ values can be used to identify AIGTs. In this section, we provide evidence that supports the assumption by comparing the trend of $p$ and $r$ values.

Table \ref{tab:pr_diff} on page \pageref{tab:pr_diff} illustrates the average of precision and recall values we collected. On average, the table shows that $p_2$ and $r_2$ are smaller than $p_1$ and $r_1$, respectively. This indicates that $T_2$ was semantically far from $T_0$ than $T_1$. So, as we apply rephraser more times on $T_0$, the semantics of rephrased text becomes farther from $T_0$.

Also, the result shows that $p$ and $r$ values are lower in human-written texts than AIGTs. For example, human-written text showed $p_1$ of 0.619, which is lower than AIGTs (ranging from 0.636 to 0.651). So, it is reasonable to use these values to distinguish between human-written texts and AIGTs.




\subsection{Effect of prompt and domain changes}
\label{app:prom}
Since we used different prompting methods and datasets in generating AIGTs, we conducted the per-subset experiment to investigate whether those differences affected the performance of detectors.
Specifically, we conducted multi-candidate experiments for each subset. For example, instead of using all data, we trained and tested models only with texts from PubMedQA.


Table \ref{tab:diff_data} on page \pageref{tab:diff_data} shows the results of the per-subset experiment. Though the domains and prompting methods are different across those subsets, \dartdt\ achieved consistently high-performance scores by showing 98.6\% macro F1. Also, \dartsvm\ (ranging from 80.0\% to 93.2\%) showed better consistency than SeqXGPT (ranging from 57.0\% to 75.1\%). This result indicates that \dart\ models are robust on changes of domains or prompting methods compared to SeqXGPT.

\subsection{Principal components of features}
\label{app:pca}

\begin{figure*}[p]
  \centering
  \includegraphics[width=1\textwidth]{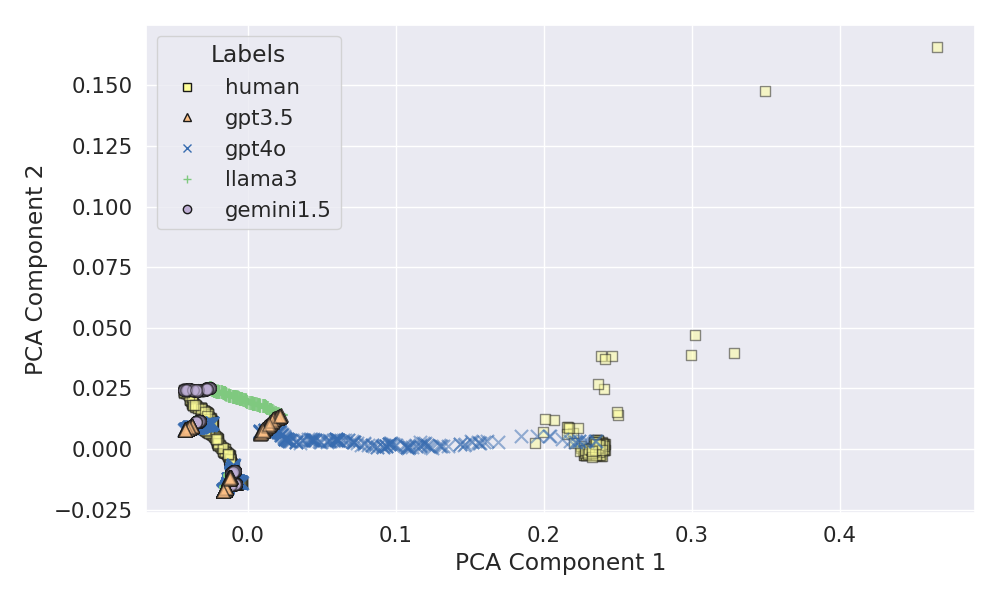}
  \caption{PCA Plot between the first principal component and the second}
  \label{fig:PCA1}
\end{figure*}

\begin{figure*}[p]
  \centering
  \includegraphics[width=1\textwidth]{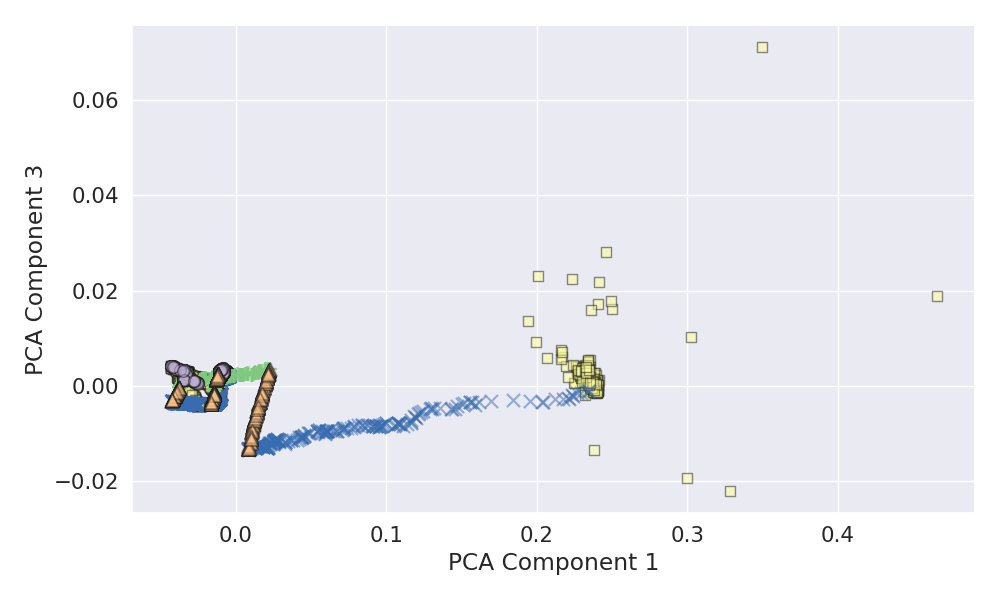}
  \caption{PCA Plot between the first principal component and the third}
  \label{fig:PCA2}
\end{figure*}

Figure \ref{fig:PCA1} and \ref{fig:PCA2} in page \pageref{fig:PCA1} display PCA plots of features used in DART. The figures show that each source makes several clusters. Here, we attempt to interpret DART's experimental results by analyzing the PCA results.
The distribution of feature vectors may affect the performance of SVM and DT classifiers. As SVM seeks a global decision boundary that maximizes margin, SVM may not find a clear decision boundary between multiple mini clusters. Meanwhile, DT can split such mini clusters based on multiple criteria. So, DT could achieve high performance in discriminating AIGTs from human-written texts. For example, we can easily discriminate humans from others and iteratively build different decision boundaries between smaller clusters. As a result, \dartdt\ can clearly discriminate sources and showed higher performance than \dartsvm.



\subsection{Training efficiency on single-candidate setting} 
\label{app:efficiency}
\begin{figure*}[t]%
\centering
\subfloat[GPT-3.5-turbo]{{\includegraphics[width=0.5\textwidth ]{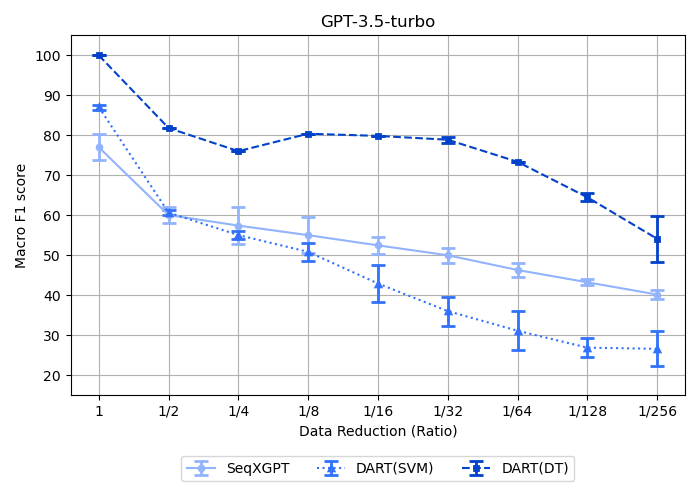} }}%
\subfloat[GPT-4o]{{\includegraphics[width=0.5\textwidth ]{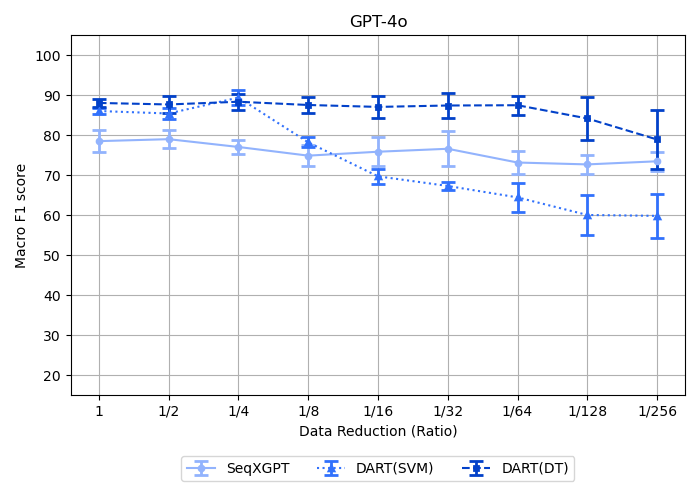} }}%
\hfill 
\subfloat[Llama 3-70b]{{\includegraphics[width=0.5\textwidth ]{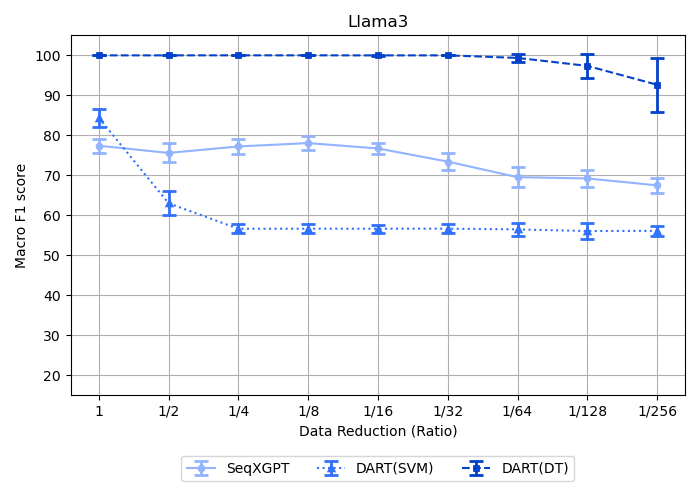} }}%
\subfloat[Gemini-1.5-Flash]{{\includegraphics[width=0.5\textwidth ]{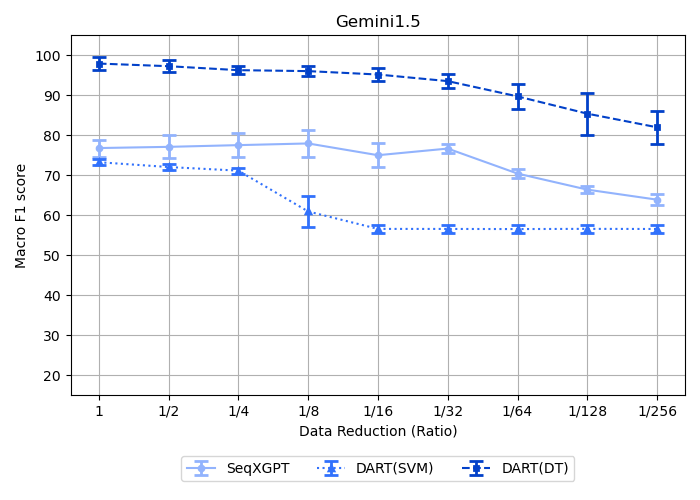} }}%
\caption{Training efficiency on the single-candidate experiment}%
\label{fig:AblationResult2}%
\end{figure*}

Figure \ref{fig:AblationResult2} in page \pageref{fig:AblationResult2} shows the training efficiency on the single-candidate experiment. In general, the performance drops as the size of the dataset decreases. Among those models, \dartdt\ demonstrates the best performance across all models, even with small datasets. \dartsvm\ experiences a more rapid decrease in its performance.



We suspect that the distribution of the data may affect the classification performance. In other words, SVM or a neural network may not have sufficient data to distinguish small clusters whose features are close to each other when we use a small dataset.


\begin{table}[!t] 
\centering
\begin{tabular}{@{}@{\;\;}lrrl}
\toprule
        & \multicolumn{1}{l}{\# char}    & \multicolumn{1}{l}{\#    tokens} & \# sample                \\
\midrule       
\multicolumn{4}{l}{PubMedQA dataset} \\
\midrule
Human         & 265.9                              & 41.8                            & \multicolumn{1}{r}{995}  \\
LLMs     & 1132.4                             & 188.1                           & \multicolumn{1}{r}{3980} \\
\quad GPT-3.5T            & 496.2                              & 78.2                            & \multicolumn{1}{r}{995}  \\
\quad GPT-4o            & 1181.4                             & 192.6                           &                          \\
\quad Llama 3-70B            & 1327.5                             & 212.7                           &                          \\
\quad Gemini-1.5F         & 1524.7                             & 268.9                           &                          \\
\midrule
\multicolumn{4}{l}{Xsum dataset} \\
\midrule
Human         & 2194.5                             & 428.9                           & \multicolumn{1}{r}{999}  \\
LLMs       & 909.5                              & 160.5                           & \multicolumn{1}{r}{3996} \\
\quad GPT-3.5T            & 773.7                              & 136.5                           & \multicolumn{1}{r}{999}  \\
\quad GPT-4o            & 1627.8                             & 282.4                           &                          \\
\quad Llama 3-70B            & 671.9                              & 121.6                           &                          \\
\quad Gemini-1.5F         & 564.7                              & 101.5                           &                          \\
\midrule
\multicolumn{4}{l}{Reddit dataset} \\
\midrule
Human         & 2962.7                             & 641.0                           & \multicolumn{1}{r}{997}  \\
LLMs          & 1135.5                             & 237.3                           & \multicolumn{1}{r}{3988} \\
\quad GPT-3.5T            & 852.3                              & 176.7                           & \multicolumn{1}{r}{997}  \\
\quad GPT-4o            & 1986.7                             & 413.5                           &                          \\
\quad Llama 3-70B            & 1009.5                             & 213.0                           &                          \\
\quad Gemini-1.5F         & 691.4                              & 146.0                           &                          \\
\midrule
\multicolumn{4}{l}{SQuAD dataset} \\
\midrule
Human         & 740.2                              & 135.1                           & \multicolumn{1}{r}{998}  \\
LLMs       & 947.5                              & 157.0                           & \multicolumn{1}{r}{3992} \\
\quad GPT-3.5T            & 503.4                              & 79.1                            & \multicolumn{1}{r}{998}  \\
\quad GPT-4o            & 1803.1                             & 303.6                           &                          \\
\quad Llama 3-70B            & 809.7                              & 142.4                           &                          \\
\quad Gemini-1.5F         & 673.9                              & 102.8                           &                          \\
\bottomrule
\end{tabular}
\caption{Statistics of collected datasets}
\label{tab:statistics}
\end{table}

  
  
  
  

\end{document}